\documentclass[conference]{IEEEtran}
\IEEEoverridecommandlockouts
\usepackage{cite}
\usepackage{amsmath,amssymb,amsfonts}
\usepackage{algorithmic}
\usepackage{graphicx}
\usepackage{textcomp}
\usepackage{xcolor}
\usepackage{booktabs}
\usepackage{subcaption}
\usepackage{comment}
\usepackage{float}
\usepackage{flushend}
\usepackage{url}
\usepackage{xcolor}

\usepackage{hyperref}

\pdfoutput=1

\def\BibTeX{{\rm B\kern-.05em{\sc i\kern-.025em b}\kern-.08em
    T\kern-.1667em\lower.7ex\hbox{E}\kern-.125emX}}
\begin{document}

\title{Hierarchical Policy Learning for Mechanical Search
{\footnotesize }
\thanks{}
}

\author{Oussama Zenkri$^{1}$, Ngo Anh Vien$^{2}$, Gerhard Neumann$^{1}$
\thanks{$^{1}$ Karlsruhe Institute of Technology (KIT)}%
\thanks{$^{2}$ Bosch Center for Artificial Intelligence (BCAI)}%
}

\maketitle

\begin{abstract}
Retrieving objects from clutters is a complex task, which requires multiple interactions with the environment until the target object can be extracted. These interactions involve executing action primitives like grasping or pushing as well as setting priorities for the objects to manipulate and the actions to execute. Mechanical Search (MS) \cite{MS} is a framework for object retrieval, which uses a heuristic algorithm for pushing and rule-based algorithms for high-level planning. While rule-based policies profit from human intuition in how they work, they usually perform sub-optimally in many cases. Deep reinforcement learning (RL) has shown great performance in complex tasks such as taking decisions through evaluating pixels, which makes it suitable for training policies in the context of object-retrieval. In this work, we first formulate the MS problem in a principled formulation as a hierarchical POMDP. Based on this formulation, we propose a hierarchical policy learning approach for the MS problem. For demonstration, we present two main parameterized sub-policies: a push policy and an action selection policy. When integrated into the hierarchical POMDP's policy, our proposed sub-policies increase the success rate of retrieving the target object from less than 32\% to nearly 80\%, while reducing the computation time for push actions from multiple seconds to less than 10 milliseconds. 
\end{abstract}
Video: \href{https://youtu.be/cioawhgFiLU}{\textit{\color{blue}\underline{\smash{{youtu.be/cioawhgFiLU}}}}} 

\begin{IEEEkeywords}
Mechanical search, action planner, push policy, hierarchical policy.
\end{IEEEkeywords}

\section{Introduction}

In recent years, the use of robots in various fields has seen an unprecedented increase. The use cases of robots, however, remained almost unchanged. This is due to the limitation of robots being designed to mainly operate in structured environments. Instead of robots adapting to their environments, work environments are usually adapted to robots. These inconveniences increase their procurement costs and limit their fields of application. In an unstructured environment, such as warehouses or homes, simple actions like grasping objects present a very demanding task for robots, which fails in many cases \cite{katz2008can}. Although robots are much faster, much more precise, and much stronger than humans, they are significantly outperformed by humans when it comes to manipulation tasks. Humans show this superior performance thanks to the disposal of a nearly endless repertoire of motion primitives that they can easily perform, as well as the ability to safely predict the outcome of any motion primitive \cite{dogar2011framework}, which robots lack.

In this paper, we want to learn how to grasp a specific object from a heap of cluttered objects. The target object might not be visible in the beginning and other objects might need to be removed or pushed away to make the target object visible and reachable. We will base our work on an existing framework from a recent work called Mechanical Search \cite{MS}, where the authors combine different algorithms for object recognition and learning grasp points. For object recognition, SD Mask-RCNN \cite{danielczuk2019segmenting} and a Siamese network \cite{Koch2015SiameseNN} are used, while Dex-Net \cite{mahler2017dex2} is used for selecting the grasp point. In this approach, the objects to be removed are currently selected via heuristics. 

In this work, we first formulate the MS problem in a principled formulation as a hierarchical POMDP. Based on this formulation, we propose a hierarchical policy learning approach. Policies higher in the hierarchy, e.g. action selection policies, will be optimized through the use of hierarchical dynamic programming which carries Bellman updates based on the value functions of policies lower in the hierarchy (called sub-policies), e.g. push and grasp policies. Policies at each layer in the hierarchy can be optimized in bottom-up fashion \cite{pineau} or jointly \cite{kulkarni2016hierarchical}. For demonstration, we present two main parameterized sub-policies: a push policy and an action selection policy. We propose to use reinforcement learning to optimize these policies. When integrated into the global hierarchical policy to do MS, our proposed solution increases the success rate of retrieving the target object in a clutter of 20 objects from less than 32\% to nearly 80\%.  
\section{Related Work}

Many research projects, which deal with developing the cognitive abilities of robots have made significant improvements in the fields of perception, planning, and control. However, less progress has been made on more complex tasks, which require a combination of these fields, like retrieving objects in unstructured environments, such as warehouses or offices. 

An early work by Li et al. \cite{7759839} demonstrates how modelling the problem of object search in clutter as a POMDP improves action planning and thus reduces the number of required steps to retrieve a target object. However, this work mainly deals with the problem of partial occlusion. 

Danielczuk et al. \cite{MS} propose mechanical search that can deal with extracting objects from clutter. 
Mechanical search uses SD Mask R-CNN \cite{danielczuk2019segmenting}, a depth-based category-agnostic segmentation algorithm, to distinguish objects and a Siamese network \cite{Koch2015SiameseNN} to identify the target object. It uses Dex-Net \cite{mahler2017dex2}, \cite{mahler2018dex3} to plan grasps and a heuristic policy to plan pushes \cite{LinPush}. All action policies produce a quality value for the action they plan.
Which action is to apply on which object is also decided heuristically. The target object, if found, has always the highest priority for manipulation. If not found or impossible to apply actions on the target object, the other detected objects are sorted in descendent order according to their visible area.
In a second work, Danielczuk et al. \cite{DanielczukAVG20} propose a follow-up idea to improve the search by estimating the occupancy distribution of the target object, which is often occluded from the scene. A further follow-up work from the same authors tries to train a push policy \cite{kurenkov2020visuomotor} via learning from demonstration and RL, which shares a similar direction to our proposed approach. However, we resort to motion primitives with task parameterization to achieve sample efficiency instead of using teacher guidance that is not always available. 

Sarantopoulos et al. \cite{modularrl21} introduce a framework for singulating objects through pushing. It incorporates two push primitives, which evaluate a latent representation of the visual state to generate push candidates for the target object and the obstacles. The high-level policy decides which push to apply. While singulation can speed up object retrieval, this approach is only beneficial in scenes  with enough free space.
Another related work by Dogar et al. \cite{dogar2011framework} which also deals with extracting objects from clutter presents a very similar approach. It uses a repertoire of human-inspired primitives like pushing, grasping, end effector motions, and combinations. 
Novkovic et al. \cite{9197101} introduces a framework for uncovering objects in cluttered scenes through interactive perception. The used policy processes an encoded volumetric truncated signed distance field (TSDF) representation of the RGB-D observation to generate some end effector displacement.
Pan et al. \cite{pan2020decision} introduce a bi-level motion planner, which deals with making optimal decisions in a joint push-grasp action space for object sorting. 
Zeng et al. \cite{zeng2018learning} present a new method for planning pushes and grasps using two end-to-end deep networks to capture complementary pushing and grasping policies that benefit from each other via reinforcement learning. 
Another work by Deng et al. \cite{8967899} propose a deep RL approach for pushing and picking in clutter. 
The RGB-D observation is transformed into an affordance map, which is used to determine if a grasp is possible and where to perform it. The agent disturbs the scene by performing a push action if grasping fails or if no grasping is possible.
Berscheid et al. \cite{berscheid2019robot} and Feldman et. al. \cite{bcai} present a vision-based algorithm for learning the most rewarding pose for applying object manipulation primitives. 
These algorithms are targeted to learn a policy that chooses between shifting and grasping actions in a bin-picking setting. However these methods are not designed for searching for a target object.


\section{Problem Formulation}
\label{problem}
Different from the original formulation from Danielczuk et al. \cite{MS}, we formulate the object search problem as a hierarchical POMDP \cite{pineau,VienT15}. Similar to a standard POMDP as defined by Danielczuk et al. \cite{MS}, a hierarchical POMDP is defined as a tuple ($\mathcal{S}, \mathcal{A}, \mathcal{T}, \mathcal{O}, \mathcal{R}, \mathcal{Y}$).

\begin{itemize}
    \item \textbf{State} $s\in \cal S$: a set of ground-truth information of the environment, i.e. all object's geometry and pose, robot's states, all sensor's states.
    \item \textbf{Action} $a \in \mathcal{A}$: a set of robot commands, e.g. a next end effector (EE) pose.
    \item \textbf{Observation} $y \in \mathcal{Y}$: a set of sensor data, which is an RGB-D image in our setting.
    \item \textbf{Transition} $\cal T$: defines a state transition given action $a$, i.e. ${\cal T}:{\cal S} \times {\cal S} \times {\cal A}\mapsto [0,1]$ where ${\cal T}(s',s,a) =p(s'|s,a)$. $s'$ denotes the following state of $s$.
    \item \textbf{Observation function} $\mathcal{O}$: defines an observation distribution given ground-truth state $s$, i.e. $\mathcal{O}(y,s)=p(y|s)$.
    \item \textbf{Reward function} $\mathcal{R}$:  defines a reward given state $s$ and action $a$, i.e. $R(s,a)$.
    \end{itemize}
Finding a global policy for the above POMDP problem is hard \cite{pineau}. In literature, there has been much work attempting to reduce problem complexity by proposing hierarchical approaches or action decomposition.

In principle, a hierarchical POMDP can be decomposed as a hierarchy of partially observable semi-MDPs (POSMDPs) \cite{white1976procedures}. In particular, assuming a hierarchical POMDP can be solved with a hierarchical policy $\pi=\{\pi_0,\pi_1,\ldots,\pi_k\}$, where $\pi_0,\pi_1,\ldots,\pi_k$ are called sub-policy, macro action, parameterized motion primitives, etc. in different contexts. Each sub-policy is targeted to solve a sub-POMDP problem (sub-task) which has much lower complexity than the global task. This principled policy decomposition can result in hierarchical Bellman updates, hence hint a principled way to optimize a hierarchically or recursively optimal policy (the type of optimality would depend on how each sub-task's reward function is defined), i.e. via hierarchical dynamic programming.

In regard to the approach of Danielczuk et al. \cite{MS}, the authors have also come up with a hierarchical policy design. The global policy $\pi$ is decomposed into a search policy $\pi_0$ whose only child action is the object selection policy $\pi_1$. Policy $\pi_1$'s child actions are segmented objects $\{\pi_2,\ldots,\pi_{N+2}\}$ (assuming there are $N$ objects in the scene at a particular time $t$). Each policy $\pi_i$ ($2\le i \le N+2$) has the same action policy that contains a set of two actions: grasp action policy $\pi_{N+3}$ and push action policy $\pi_{N+4}$. Each of the grasp or push action policies are parameterized motion primitives that would use low-level action $a\in \cal A$ as primitive actions. However, Danielczuk et al. \cite{MS} have not yet discussed if each sub-policy $\pi_i$ is derived from solving a well-defined sub-POMDP problem. Among them, grasp action policy $\pi_{N+3}$ is a pretrained Dex-Net policy network \cite{mahler2017dex2} which was shown to come from a POMDP grasping problem. The other sub-policies are heuristic and hard-coded, therefore not designed as a solution to a well-defined sub-POMDP problem. Thus, this approach could have a great practical benefit but might lack the flexibility of extension to more powerful solutions, e.g. using hierarchical RL.

Based on our above hierarchical POMDP formulation and action decomposition, we propose two extensions that transform i) the action selection problem to a POMDP problem with a trainable \emph{action selection policy} (depending on its child actions), and ii) the push action to a POMDP problem with a trainable \emph{push policy}. Our following proposals assume separate training for simplicity, however, it can also be trained in an end-to-end hierarchical POMDP RL fashion \cite{kulkarni2016hierarchical}.

\section{Push Policy}
\label{sec:push_policy}

Danielczuk et al. \cite{MS} use the free space policy (FSP) \cite{LinPush}, as a push primitive in the MS pipeline. The FSP aims to push the target object (TO) to the freest space in the bin. This would increase the free space around the TO and consequently increases the probability of a successful grasp. Danielczuk et al. conjuncture that this policy performs suboptimal and suggest replacing it with more effective push primitives, which can learn from simulation. Our observations validate this assumption. While the FSP performs well in small heaps, it shows poor performance in cluttered scenes where increasing the free space around an object is difficult. Moreover, the FSP fails at inferring a push action for many observations and takes multiple seconds to compute one push action. These drawbacks motivate us to develop a new push policy, which learns pushing from simulation. First, we introduce a problem statement for the linear pushing, then we introduce our method for solving the push problem. 

\subsection{Problem Statement}
\label{sec:push_prob_statement}

We formulate the general linear push problem as a sub-POMDP defined by ($\mathcal{S}, \mathcal{A}, \mathcal{T}, \mathcal{R}, \mathcal{Y}$), where it shares the state space $\cal S$ of the main problem defined in Section \ref{problem}. For this problem we precisely define the observations, the action set, and the reward function. All other aspects of the problem formulation are sufficiently captured by the main POMDP.

\begin{figure}[h]
\vspace{-2mm}
     \centering
     \begin{subfigure}[b]{0.11\textwidth}
         \centering
         \includegraphics[width=\textwidth]{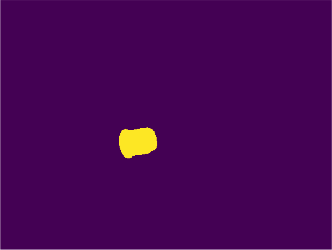}
         \caption{}
         \label{fig:rl_push_reward_mask}
     \end{subfigure}
     \hfill
     \begin{subfigure}[b]{0.11\textwidth}
         \centering
         \includegraphics[width=\textwidth]{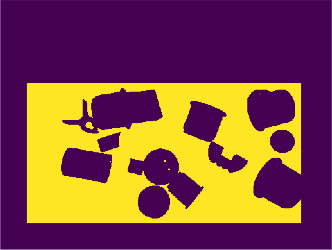}
         \caption{}
         \label{fig:rl_push_reward_bb}
     \end{subfigure}
     \hfill
     \begin{subfigure}[b]{0.11\textwidth}
         \centering
         \includegraphics[width=\textwidth]{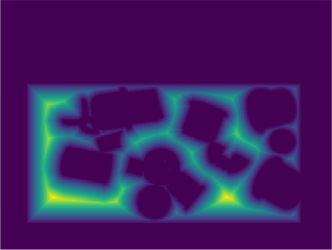}
         \caption{}
         \label{fig:rl_push_reward_dist}
     \end{subfigure}
     \hfill
     \begin{subfigure}[b]{0.11\textwidth}
         \centering
         \includegraphics[width=\textwidth]{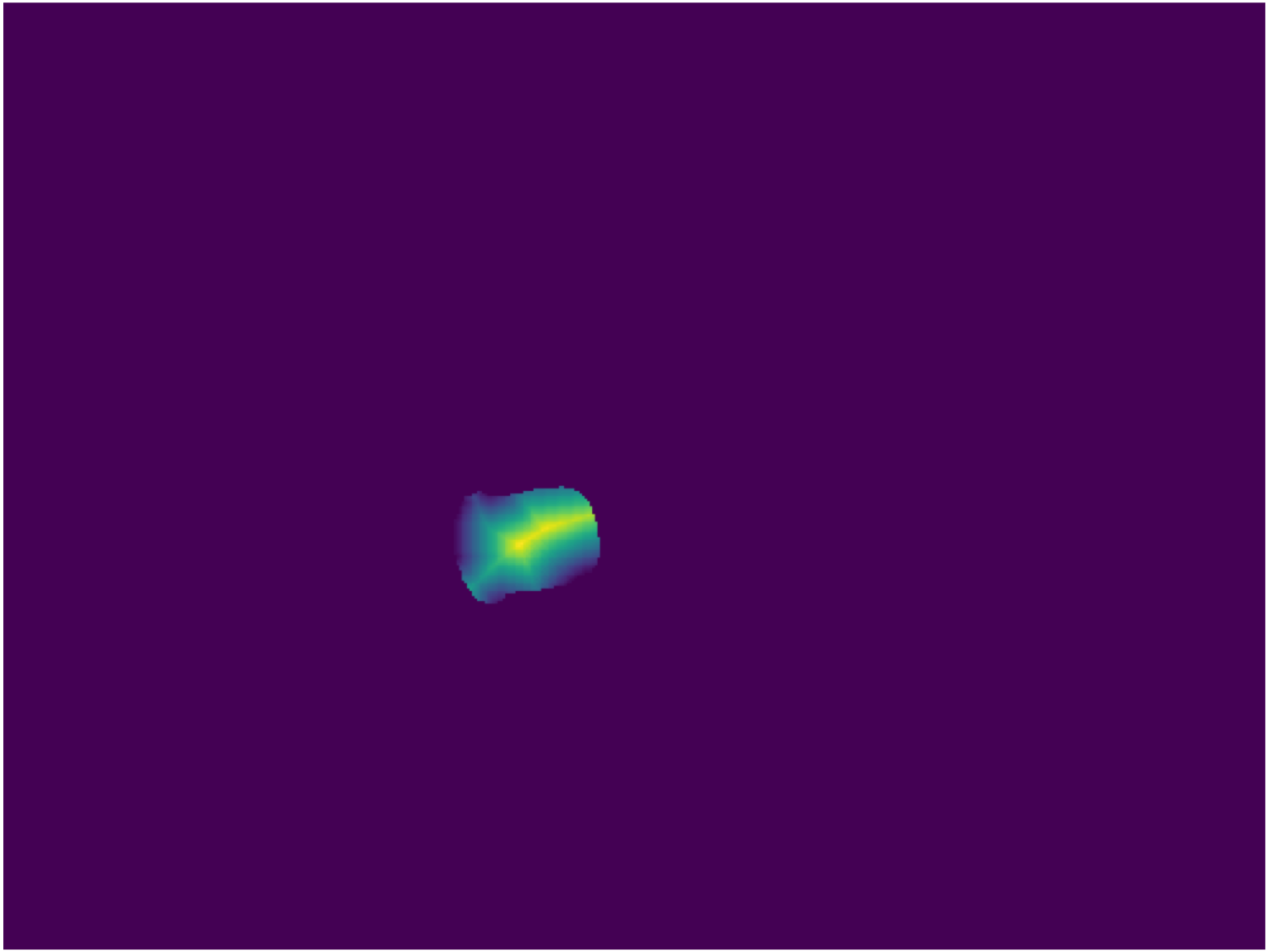}
         \caption{}
         \label{fig:rl_push_reward_final}
     \end{subfigure}
        \caption{Elements of the computation of the free space for object $O_i$. (a) Binary mask of object $O_i$. (b) Bin free space mask. (c) Distance transform of $\mathbf{M}_{\text{BFS}, O_i}$. (d) Masked $\mathbf{DT}_{\text{BFS}, O_i}$.}
        \label{fig:rl_push_reward_steps}

\end{figure}

\begin{itemize}
    \item \textbf{Observation} $y \in \mathcal{Y}$: a downscaled crop from the depth image centered around the target object. The crop is sized $220\times220$ (px) and downscaled to $40\times40$ (px).
    \item \textbf{Action} $a \in \mathcal{A}$: a six-dimensional continuous action defined by $x_{rel}$, $y_{rel}$, $\operatorname{sin}\alpha_{\text{Push}}$, $\operatorname{cos}\alpha_{\text{Push}}$, $\operatorname{sin}\phi$, $\operatorname{cos}\phi \in [-1,\,1]$. The action encodes the relative position of the EE at push start $(x_{rel}, y_{rel})$, the push direction $\alpha_{\text{Push}}$ and the yaw angle of the EE $\phi$. Values $x_{rel}$ and $y_{rel}$ are relative to the observation image. 
    The absolute coordinates in the depth image $(u, v)$ can be deduced from the size and coordinates of the crop. The first four variables of the action space are visualized in Fig. \ref{fig:rl_push_action_space}.
    The push start position $\mathbf{p}_{\text{Start}}$ is obtained through deprojecting $u$, $v$ and the estimated depth $z$.
    The push distance is $10$ cm. The push end point $\mathbf{p}_{\text{End}}$ is inferred from $\mathbf{p}_{\text{Start}}$ and $\alpha_{\text{Push}}$.

    \begin{figure}[h]
    \begin{center}
    \includegraphics[width=0.2\textwidth]{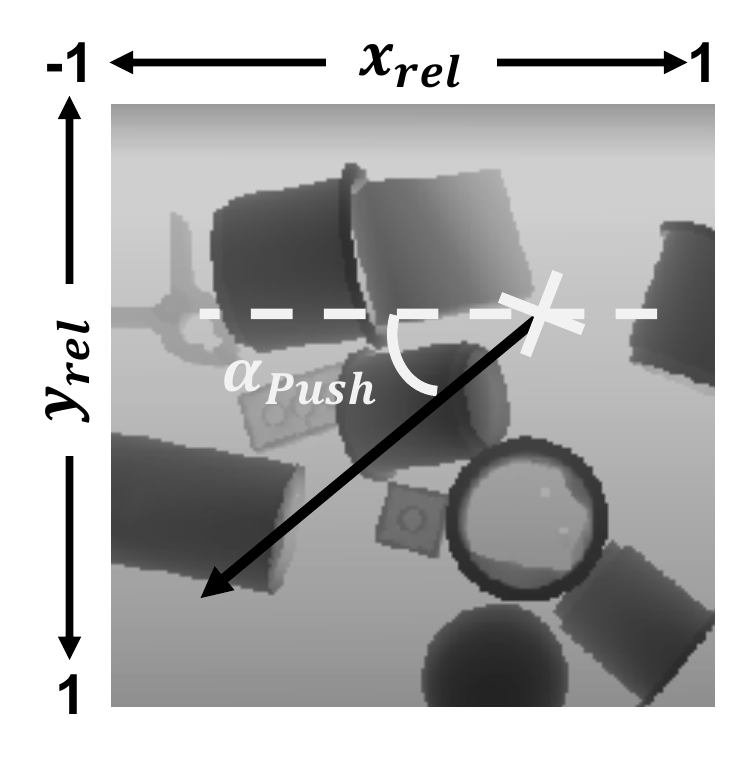}
    \caption{Visualization of the action space variables for the push policy. The position of EE at push start $(x_{rel}, y_{rel})$ is shown as a white cross. The push direction is defined by $\alpha_{\text{Push}}$. The EE-yaw angle $\phi$ is not represented in this figure. 
    }
    \label{fig:rl_push_action_space}
    \end{center}
    \vspace{-4mm}
    \end{figure}
    
    \item \textbf{Reward} $r \in \mathcal{R}$: a value quantifying the change in free space around the OOI. The reward function is defined as the weighted sum of changes in free space around each object present in the bin:
    \begin{equation}
        r_t = \frac{10\cdot\Delta_{\text{FS}, O_I}(t) + \frac{1}{N-1}\cdot\sum_{i = 1, i \neq I}^{N} \Delta_{\text{FS}, O_i}(t)}{11}
    \end{equation}
    where, $N$ denotes the number of objects in the bin. $O_I$ denotes the OOI. $\Delta_{\text{FS}, O_i}$ denotes the change in free space (FS) around object $O_i$ after executing the push action as denoted in Eq. \ref{eq:delta_fs}. We reward the increase of free space around all objects, while focusing on the OOI.
    \begin{equation}
        \Delta_{\text{FS}, O_i}(t) = \text{FS}_{O_i}(t) - \text{FS}_{O_i}(t-1).
        \label{eq:delta_fs}
    \end{equation}
    Here, $\text{FS}_{O_i}(t)$ denotes the free space around $O_i$ at time $t$ and is computed as the sum of the masked distance transform of the bin free space for $O_i$ (Fig. \ref{fig:rl_push_reward_final}), normalized by the area of $O_i$  
    \begin{equation}
        \text{FS}_{O_i}(t) = \frac{\sum{(\textbf{M}_{O_i}(t) \odot \mathbf{DT}_{\text{BFS}, O_i}(t)})}{\sum\textbf{M}_{O_i}(t)}
        \label{eq:free_space}
    \end{equation}
    where, $\mathbf{DT}_{\text{BFS}, O_i}$ denotes the result of the distance transform of the bin free space (BFS) mask for object $O_i$ (Fig. \ref{fig:rl_push_reward_dist}). $\sum\textbf{M}_{O_i}$ is a scalar denoting the area of the binary mask $\textbf{M}_{O_i}$ of object $O_i$ (Fig. \ref{fig:rl_push_reward_mask}). The operator $\odot$ denotes the element-wise multiplication. The BFS-mask (Fig. \ref{fig:rl_push_reward_bb}) results from the subtraction of all observed object masks from the bin bottom (BB) mask except for the OOI:
    \begin{equation}
        \mathbf{M}_{\text{BFS}, O_i}(t) = \mathbf{M}_{\text{BB}} - \sum_{i = 1, i \neq I}^{N} \mathbf{M}_{O_i}(t)
    \end{equation}

\end{itemize}

\subsection{Push Policy Learning}
\label{sec:push_method}

\begin{figure}[ht]
\begin{center}
\includegraphics[width=0.45\textwidth]{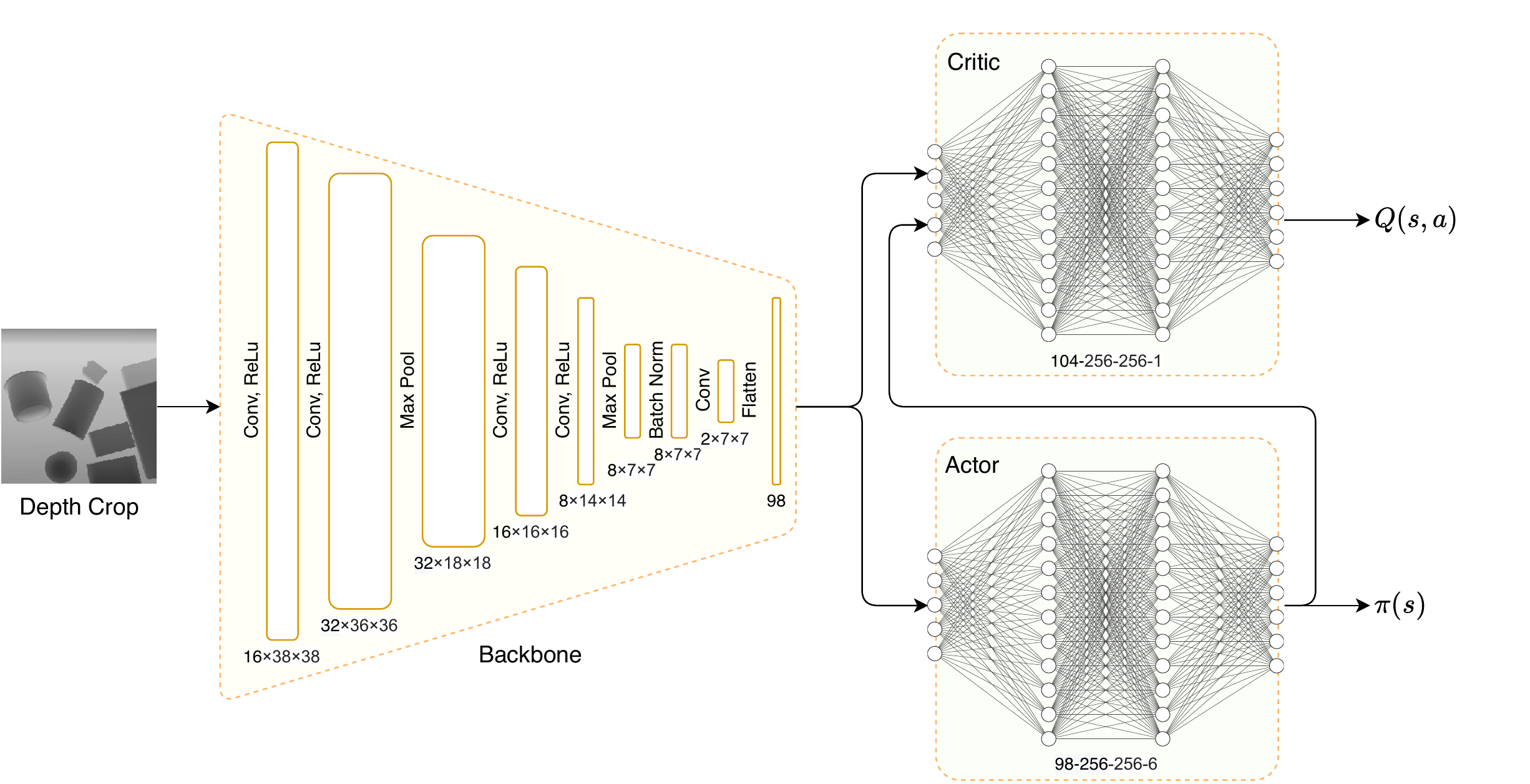}
\caption[Architecture of our push policy]{Architecture of our push policy. The network consists of a CNN encoder for feature extraction, a fully connected actor-network serving as policy, and a fully connected critic-network, which serves as an action-value function.}
\label{fig:rl_push_policy}
\end{center}
\vspace{-4mm}
\end{figure}

For our push policy we use an actor-critic architecture, which consists of two branches: an actor-network and a critic-network. The actor takes a state observation as input and returns an action according to a policy $\pi_\theta$. The critic serves as an action-value function and returns a quality value for the computed action given the observed state. These two networks are optimized jointly during the training. Our architecture also includes an encoder, which transforms the input image of size $40\times 40$ (px) to a feature vector of size $98$. This feature vector serves as the state observation for the actor and the critic. We train the encoder jointly with the other networks. The architecture of our push policy is shown in Fig. \ref{fig:rl_push_policy}. The generated action vector $\mathbf{a}_t = (x_{rel,\,t},\: y_{rel,\,t},\: \operatorname{sin}\alpha_{\text{Push},\,t},\: \operatorname{cos}\alpha_{\text{Push},\,t},\: \operatorname{sin}\phi_t,\: \operatorname{cos}\phi_t)$ is concatenated with the state observation and forwarded to the critic-network. We optimize our policy using the soft actor-critic (SAC) algorithm \cite{HaarnojaZAL18}.

\section{Action Selection Policy}

The heuristic action selection policy (ASP), introduced in \cite{MS} has a simple tree structure, which prioritizes grasping over pushing. If neither actions are possible, then the current object is skipped. The policy uses static threshold values $q_{Grasp, Thresh}$ and $q_{Push, Thresh}$ to determine the priorities of the actions, which can result in suboptimal decisions, especially in cases of wrong quality estimations. This suboptimal behaviour can be observed in the physical rollouts introduced in \cite{MS}, where the heuristic policy is outperformed by humans, who require less than three actions to succeed in more than $50\%$ of the trials, while the introduced heuristic policy requires at least five actions to reach the same success rate. The superior human performance results i.a. from making better action selection decisions as shown in Table \ref{tab:original_vs_human}. The table shows a remarkable difference in the proportions of executed actions. This motivates us to develop a new ASP, which learns a better action selection strategy from trial and error. First, we formulate a problem statement for the action selection, then we introduce our method for solving this problem.

\begin{table}[ht]
\begin{center}
    \caption{Proportions of taken actions by the heuristic ASP and humans.}
    \begin{tabular}{lll}
    \toprule
     & \textbf{Heuristic ASP}$^*$ & \textbf{Human ASP}$^*$ \\
    \midrule
    \textbf{Parallel-Jaw Grasp}  & $18.4\%$ & $15.9\%$ \\
    \textbf{Suction Grasp} & $76\%$ & $58.2\%$ \\
     \textbf{Push} & $5.6\%$ & $25.9\%$ \\
    \bottomrule
    \end{tabular}
\label{tab:original_vs_human}
\end{center}
*: the values result from 50 physical rollouts. Reported in \cite{MS}
\end{table}

\subsection{Problem Statement}
\label{sec:asp_prob_statement}

The objective of our ASP is to select one of the available action primitives such that the number of required actions to extract a TO is minimized. We formulate the problem of action selection as a sub-POMDP problem that also shares the same state space as the main POMDP problem defined in Section \ref{problem}. For this sub-POMDP task (action selection), we specify the observations, the action set and the reward function.
\begin{itemize}
    \item \textbf{Observations}: the environment returns the following observation set:
    \begin{enumerate}
        \item \textbf{Depth Image}: a $40\times40$ (px) rescaled crop of the depth image centered around the OOI. The crop size is $220\times220$ (px), which captures sufficient context from the original image of size $480\times 640$ (px). 
        \item \textbf{OOI-Mask}: a $40\times40$ (px) rescaled crop of the OOI-mask centered around the OOI. The crop size is $220\times220$ (px).
        \item \textbf{Relative Coordinates}: $x_{rel}, y_{rel} \in [-1, \, 1]$ designate the relative coordinates of the TO in the original depth image. The absolute position of the TO is inferred from these values.
        \item \textbf{Action Qualities}: $q_{Grasp}, q_{Push} \in \{-1,\, [0,\,1]\}$ the quality metrics returned by the action policies. While the policies return values in $[0,\,1]$, we set the quality value to $-1$, if the policy could not compute an action for the OOI. This signals the agent that the corresponding action is infeasible. This behavior is reinforced with the used reward function (Eq. \ref{eq:rl_asp_reward_eq}). Selected infeasible actions are changed to \textit{Skip}.
    \end{enumerate}
    \item \textbf{Actions}: the agent selects one of the following actions:
    \begin{enumerate}
        \item \textbf{Skip}: the robot does not apply any action on the current object. Another object will be evaluated.
        \item \textbf{Grasp}: the robot grasps the current object. If the OOI is marked as TO, the object is dropped in a dedicated spot. Otherwise, the robot drops the object in a secondary bin. We assume the grasp policy is a pretrained Dex-Net 2.0 policy network \cite{mahler2017dex2}.
        \item \textbf{Push}: the robot pushes the OOI according to the plan returned by the push policy. This push action is an optimized push policy discussed in Section \ref{sec:push_policy}.
    \end{enumerate}
    \item \textbf{Reward}: we define the reward function as
    \vspace{8pt}
    \begin{equation}
        r_t = \left\{\begin{array}{ll}
            20 &  , \text{if the TO is successfully extracted,}\\
            -10 & , \text{if an infeasible is action selected,} \\
            -1 & , \text{Otherwise.}
        \end{array}\right.
        \label{eq:rl_asp_reward_eq}
    \end{equation}
\end{itemize}

\subsection{Method}
\label{sec:asp_method}

\begin{figure}[ht]
\begin{center}
\includegraphics[width = 0.45\textwidth]{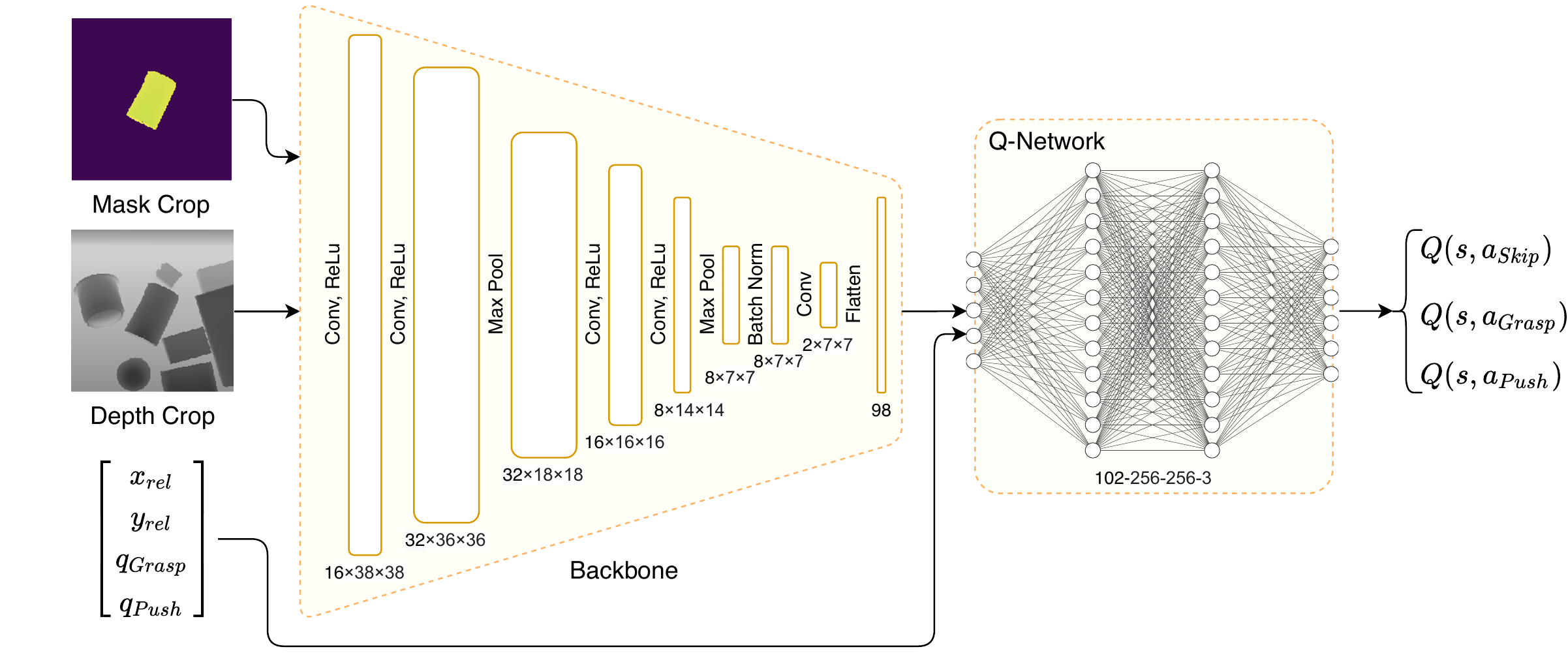}
\caption{Architecture of our ASP. The encoder extracts features from the depth image and the OOI-mask and outputs a feature vector of size $98$. The feature vector is concatenated with the scalar input data and forwarded to the $Q$-network, which outputs a quality value for each state-action pair.}
\label{fig:rl_asp}
\end{center}
\vspace{-1mm}
\end{figure}

The objective of the ASP is to select the motion primitive (grasp vs. push) to be executed, if applicable, which minimizes the number of required actions to retrieve a TO. This task has a finite discrete action space, which makes a Q-Learning algorithm a simple and effective RL method to deal with it. Thus, we use a deep Q-network (DQN) \cite{mnih2015human} architecture for our policy. 
Due to the property of the observation, which consists of a mixture of images and scalars, our network has two input branches. The first one includes an encoder for extracting features from the input images. The second branch feeds the input scalars directly in the $Q$-network.
The network outputs an action-value for each of the possible actions \{\textit{Skip}, \textit{Grasp}, \textit{Push}\}. The action with the highest $Q$-value is selected for execution. The architecture of our action selection agent is shown in Fig. \ref{fig:rl_asp}. The used encoder shares the same architecture as the one in our push policy (sec. \ref{sec:push_method}). 


\section{Experiments}
\label{sec:experiment}
\subsection{Environment}
\begin{figure}[ht]
    \centering
    \includegraphics[width=0.23\textwidth]{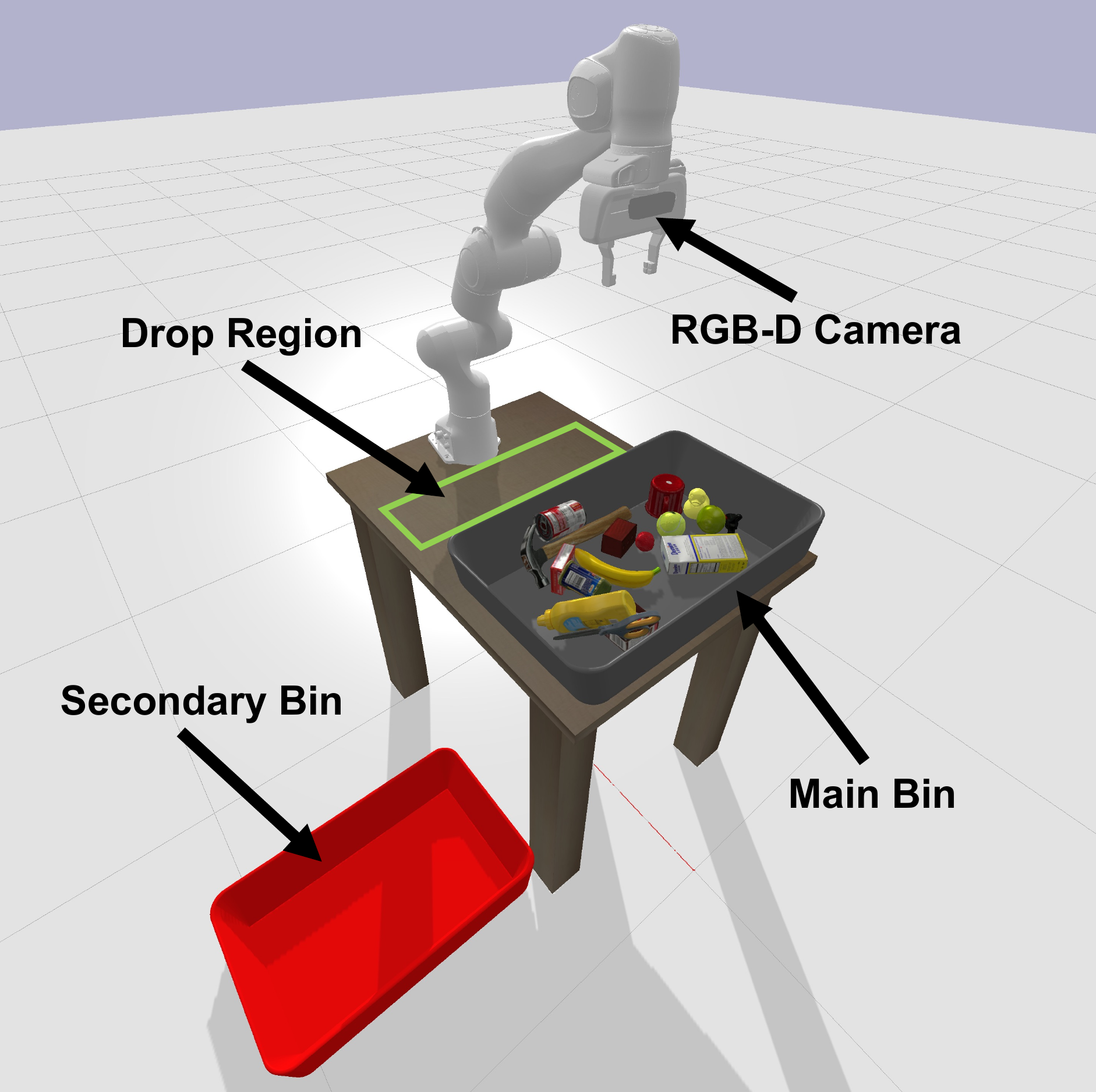}
    \caption{Overview of the work environment. The designated drop region for the TO is marked in green.}
    \label{fig:scene}
\end{figure}

In our work, we use a simulated environment running on the PyBullet physics engine \cite{coumans2017pybullet} for training our agents and benchmarking the different policies. The scene, shown in Fig. \ref{fig:scene}, consists of a simulated Franka Emika Panda robot equipped with a parallel-jaw gripper. The robot is placed on top of a table close to a bin, where the objects to manipulate are placed. A secondary bin, for dropping secondary objects, is placed beside the main bin. The used object models are a subset of the YCB dataset \cite{calli2017ycb}. The scene is perceived through a simulated Intel RealSense D-435 RGB-D camera \cite{keselman2017realsense}.

\subsection{Experiment Description}
To evaluate our policies, we compare their performances to the ones of the heuristic policies, which were introduced in the original MS pipeline \cite{MS}. 
The trials are initialized as follows:
\begin{enumerate}
    \item Randomly, select a TO from the list of available objects.
    \item Randomly, select other objects from the remaining objects, until the benchmark heap size is reached.
    \item Place the selected objects, collision free, at different locations and heights above the bin.
    \item Simultaneously, drop all objects in the bin.
    \item Allow 5 seconds for the objects to settle before making the first observation.
\end{enumerate}

After initializing the scene, the agent is allowed to sequentially apply pushes and parallel-jaw grasps to retrieve the TO. The agent can push objects in the bin or extract objects, different from the TO, and place them in the secondary bin. The trial ends when one of the stopping criteria is met:
\begin{itemize}
    \item The TO is dropped at the designated drop location.
    \item The TO is neither in the bin nor in the drop location.
    \item The number of executed actions exceeded 25.
\end{itemize}

\subsection{Training}

\subsubsection{Push Policy}

We train our push policy on clutters of ten objects, which we place randomly at different locations in the bin. We set the length of an episode to five push trials to speed up training. The agent is trained for $12$k episodes, i.e. a total of $60$k steps. The policy is updated five times after each step. The policy converges after $10$k episodes. 

\subsubsection{Action Selection Policy}

We train our ASP on heaps of size $20$. An episode is automatically terminated after $25$ steps. 
Our agent is trained for $300$ episodes. The agent is updated $20$ times after each step. The policy converges after $100$ episodes.

\subsection{Evaluated Setups}

We evaluate four different setups of the MS pipeline, which contain the four possible combinations illustrated in Table \ref{tab:setups}. All setups retain the same components of the MS pipeline, except for the push policy, and the ASP. Each setup is evaluated on two different heap sizes: $10$ and $20$ objects.

\begin{table}[h]
\begin{center}
\caption{Policy combinations of the evaluated setups}
\begin{tabular}{lcccc}
\toprule
 & \textbf{\scriptsize Heu. ASP} & \textbf{\scriptsize Our ASP} & \textbf{\scriptsize Heu. Push Pol.}$^+$ & \textbf{\scriptsize Our Push Pol.}\\
\midrule
\textbf{Setup 1} & $\times$ & & $\times$ & \\
\textbf{Setup 2} & $\times$ & & & $\times$ \\
\textbf{Setup 3} & & $\times$ & $\times$ & \\
\textbf{Setup 4} & & $\times$ & & $\times$ \\
\bottomrule
\label{tab:setups}
\end{tabular}
\end{center}
\vspace{-4mm}
+: free space policy (FSP).
\end{table}

\section{Results}

\begin{figure*}[ht]
\centering
\minipage{0.32\textwidth}
  \includegraphics[width=\linewidth]{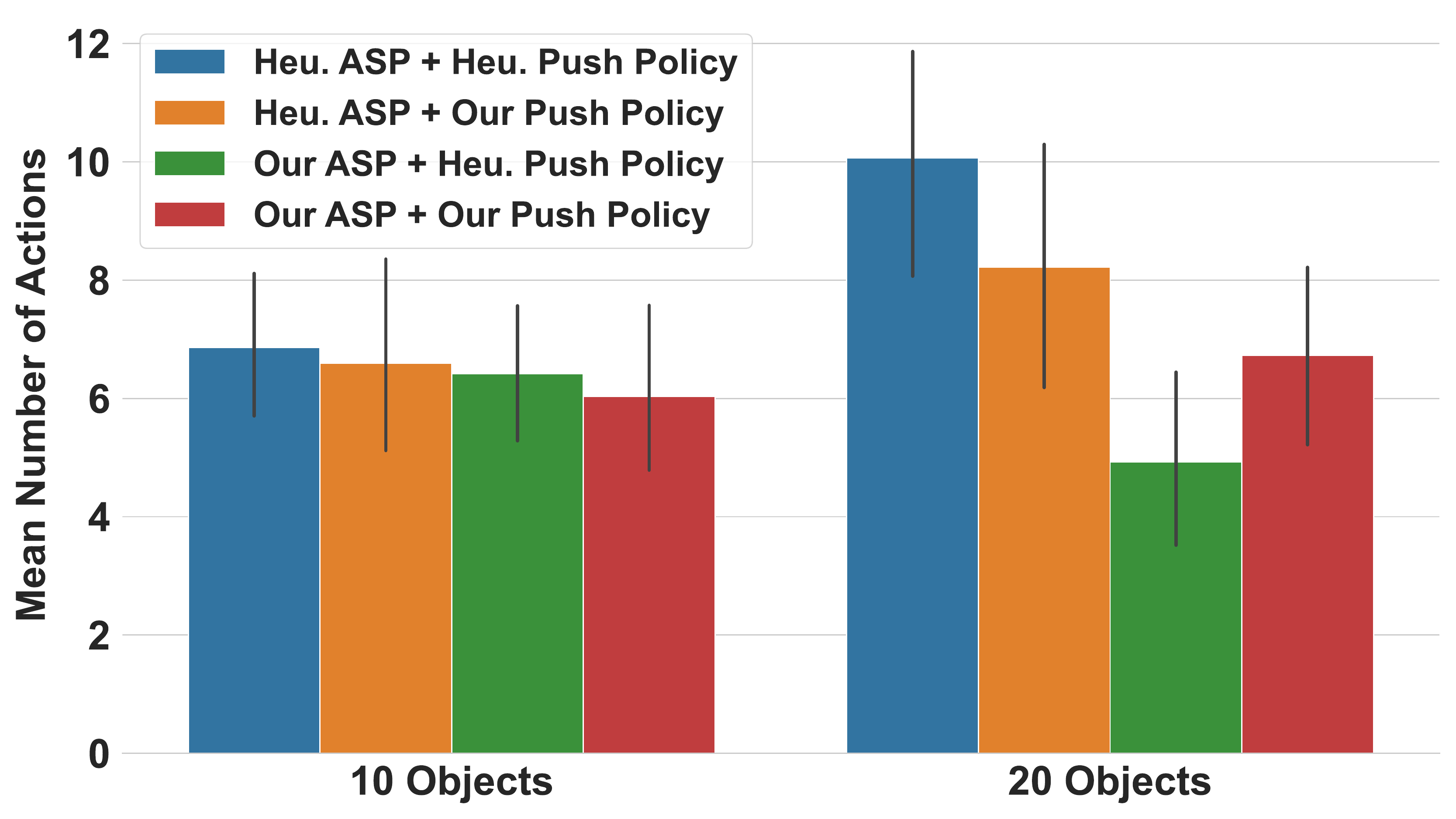}
  \subcaption{}
  \label{fig:results:a}
\endminipage\hfill
\minipage{0.32\textwidth}
  \includegraphics[width=\linewidth]{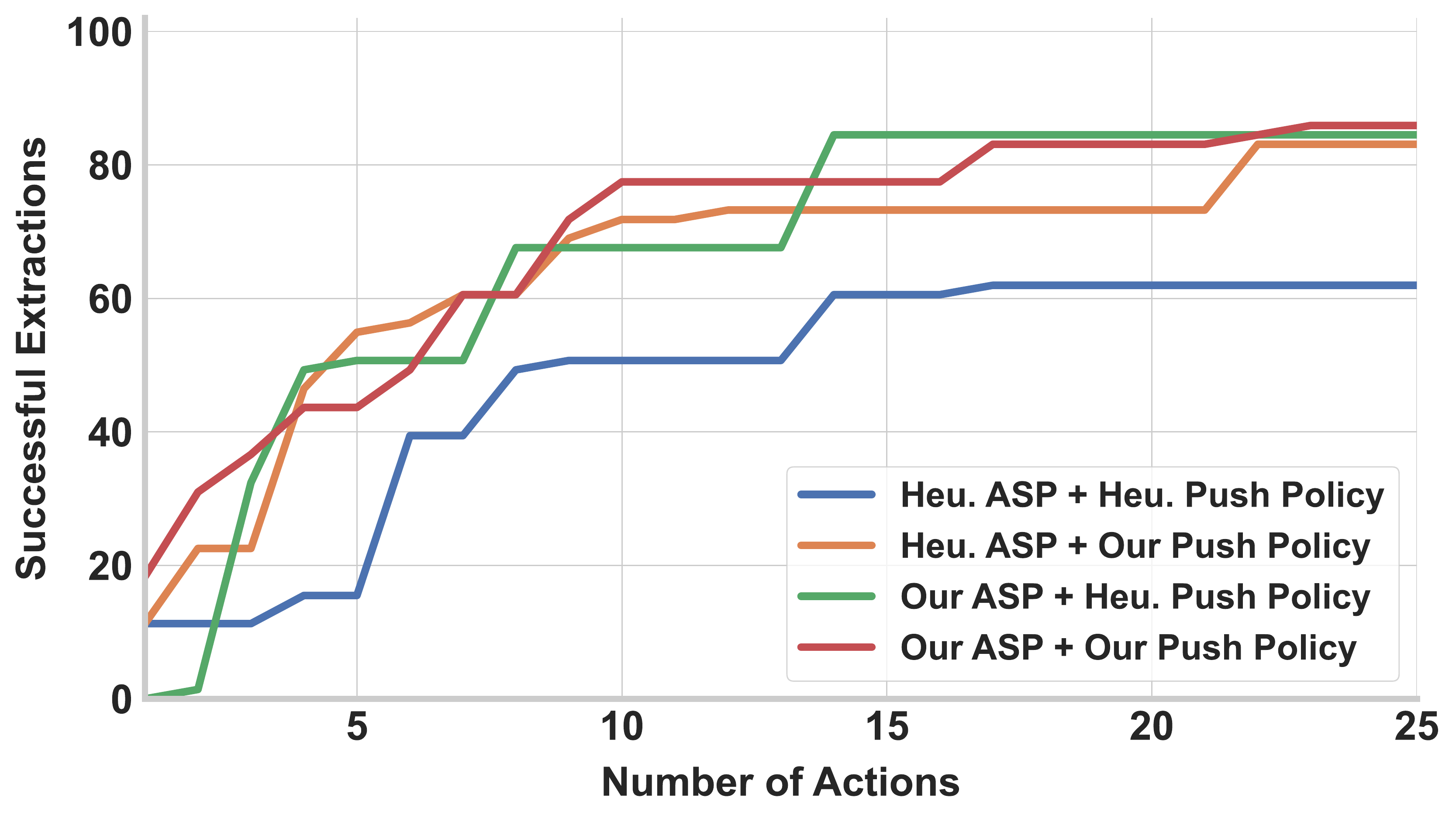}
  \subcaption{}
  \label{fig:results:b}
\endminipage\hfill
\minipage{0.32\textwidth}%
  \includegraphics[width=\linewidth]{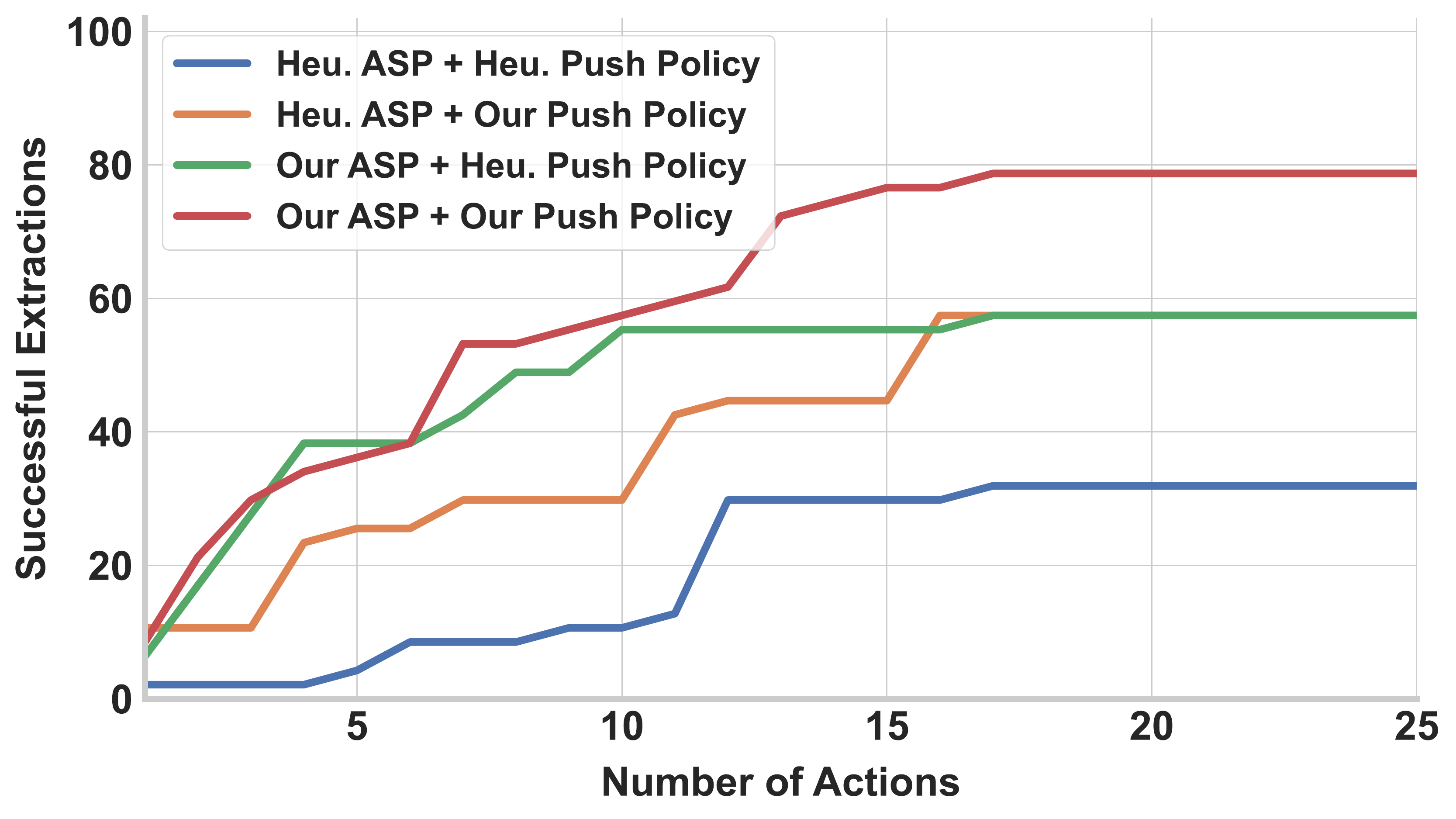}
  \subcaption{}
  \label{fig:results:c}
\endminipage
\setlength{\belowcaptionskip}{-5pt}
\caption{Performance of our policies compared to the heuristic policies of the original MS pipeline. (a) Average number of required actions for object retrieval for heaps of 10, and 20 objects. Successful extractions as function of the number of executed actions for simulated heaps of (b) size 10, and (c) size 20. 
Substituting the FSP and the ASP with our policies results in similar or better performance than swapping only one of them. This effect is more pronounced for larger heaps.
}
\label{fig:results}

\end{figure*}

We conducted $100$ simulated experiments for each policy combination and heap size resulting in $800$ rollouts. The four different policy combinations are compared against each other according to Section \ref{sec:experiment}. The results of the experiments are visualized in Fig. \ref{fig:results} and also illustrated in Table \ref{tab:results}.

Fig. \ref{fig:results:a} shows the average number of actions required by each policy combination to retrieve the TO for heaps of size $10$ and $20$. The four combinations perform similarly on the small heaps. In the larger heaps, Setup 1 performs the worst with a mean of $10.06$ actions. Setup 3 requires only $4.92$ actions. 

The success rates of the policy combinations as a function of the number of executed actions for heaps of size $10$ and $20$ are shown in Fig. \ref{fig:results:b} and Fig. \ref{fig:results:c} respectively. Both figures show that substituting the FSP by our RL-based push policy (Setup 2) leads to higher success rates and fewer required actions to retrieve the TO.
Fig. \ref{fig:results:b} shows that, when our policies are used (Setups 2, 3, and 4), the success rates exceed $40\%$ within five actions, while the success rate of the original MS pipeline (Setup 1) remains below $20\%$ for the same number of actions. If our ASP is used (Setups 3 and 4), the success rate reaches $80\%$ within $17$ actions, while the original MS pipeline settles at $60\%$. For heaps of size $20$, the success rate of the original MS pipeline (Setup 1) settles below $32\%$. If the FSP and ASP are swapped with our policies (Setup 4), the success rate reaches $79\%$ within $17$ actions.

\begin{table}[htp]
\centering
\caption{Means and standard deviations of required actions for a successful object retrieval in heaps of size $10$ and $20$}
\begin{tabular}{lllll}
\toprule
 & \textbf{Mean}$_{10}$ & \textbf{Std. Dev.}$_{10}$ & \textbf{Mean}$_{20}$ & \textbf{Std. Dev.}$_{20}$\\
\midrule
\textbf{Setup 1}  & $6.86$ & $4.22$ & $10.06$ & $3.84$ \\
\textbf{Setup 2} & $6.59$ & $6.24$ & $8.22$ & $5.46$ \\
\textbf{Setup 3} & $6.42$ & $4.22$ & $4.92$ & $3.71$ \\
\textbf{Setup 4} & $6.03$ & $5.27$ & $6.73$ & $4.72$ \\
\bottomrule
\label{tab:results}
\end{tabular}
\vspace{-2mm}
\end{table}

The comparable performance of all setups for heaps of size $10$ is due to the small size of the heap, which makes most of the objects well exposed. Consequently, the TO can be found easily. 
Setup 4 reaches higher success rates at higher numbers of actions. This biases the mean number of required actions towards higher values, thus making it appear worst than Setup 3. However figure \ref{fig:results:c} shows that Setup 4 outperforms all other setups at the majority of the given number of actions.

\begin{table}[htp]
\begin{center}
\caption{Proportions of executed actions for different ASPs}
\begin{tabular}{llll}
\toprule
 & \textbf{Original ASP} & \textbf{Human ASP}$^*$ & \textbf{Our ASP}\\
\midrule
\textbf{Parallel-Jaw Grasp}  & $86.3\%$ & $15.9\%$ & $67.9\%$ \\
\textbf{Suction Grasp} & - & $58.2\%$ & - \\
 \textbf{Push} & $13.7\%$ & $25.9\%$ & $32.1\%$ \\
\bottomrule
\label{tab:original_vs_human_vs_ours}
\end{tabular}
\end{center}
\vspace{-4mm}
*: the values result from 50 physical rollouts. Reported in \cite{MS}.
\end{table}

Table \ref{tab:original_vs_human_vs_ours} compares the proportions of executed actions by the original ASP, Humans, and our ASP. It shows that the proportion of executed pushes by the heuristic ASP is much lower than the proportion executed by humans as pointed out by Danielczuk et al. \cite{MS}. While humans make more use of pushing through selecting it more than $25\%$ of the time, less than $14\%$ of the selected actions by the heuristic ASP are pushes. Our ASP, however, shows more human-like action choice through deciding for pushing in $32\%$ of the time.  

\section{Conclusion and Future Work}
We propose a hierarchical POMDP formulation for the mechanical search problem where each sub-policy can be trainable and integrated to be optimized in a principled way. For demonstration, we introduce a new push policy, which we optimize using soft actor-critic. The aim of our push policy is to maximize the free space around a target object. We also introduce a new action selection policy, which makes use of the deep Q-network architecture to select the best action primitive to be applied for a given observation. Combined, our two policies show a significant decrease in the number of required actions to extract an object, compared to the heuristic policies introduced in \cite{MS}. This effect is more pronounced for large heaps. The success rate of object retrieval for a frame of $17$ actions increases from less than $32\%$ using the original MS pipeline to nearly $80\%$ when using our policies. Substituting the heuristic free space policy introduced in \cite{LinPush} by our push policy substantially reduces the inference time from multiple seconds to less than 10 milliseconds. Our action selection policy takes significantly more use of the push primitive than the heuristic policy in \cite{MS} and shows similar action proportions to the ones selected by humans. 

The confirmation of these results on real-world experiments, the introduction of a new learning object selection policy for ranking the objects to manipulate, and the evaluation of a joint end-to-end training of all hierarchical policies are subject of future work.

\section{Acknowledgments}

This research has been supported by HEAP (Human-Guided Learning and Benchmarking of Robotic Heap Sorting, CHIST-ERA, grant no: EP/S033718/2).

\clearpage
\bibliographystyle{IEEEtran}
\bibliography{my}

\begin{thebibliography}{10}
\providecommand{\url}[1]{#1}
\csname url@rmstyle\endcsname
\providecommand{\newblock}{\relax}
\providecommand{\bibinfo}[2]{#2}
\providecommand\BIBentrySTDinterwordspacing{\spaceskip=0pt\relax}
\providecommand\BIBentryALTinterwordstretchfactor{4}
\providecommand\BIBentryALTinterwordspacing{\spaceskip=\fontdimen2\font plus
\BIBentryALTinterwordstretchfactor\fontdimen3\font minus
  \fontdimen4\font\relax}
\providecommand\BIBforeignlanguage[2]{{%
\expandafter\ifx\csname l@#1\endcsname\relax
\typeout{** WARNING: IEEEtran.bst: No hyphenation pattern has been}%
\typeout{** loaded for the language `#1'. Using the pattern for}%
\typeout{** the default language instead.}%
\else
\language=\csname l@#1\endcsname
\fi
#2}}

\bibitem{MS}
M.~Danielczuk, A.~Kurenkov, A.~Balakrishna, M.~Matl, D.~Wang,
  R.~Mart{\'{\i}}n{-}Mart{\'{\i}}n, A.~Garg, S.~Savarese, and K.~Goldberg,
  ``Mechanical search: Multi-step retrieval of a target object occluded by
  clutter,'' in \emph{International Conference on Robotics and Automation,
  {ICRA} 2019, Montreal, QC, Canada, May 20-24, 2019}.\hskip 1em plus 0.5em
  minus 0.4em\relax {IEEE}, 2019, pp. 1614--1621.

\bibitem{katz2008can}
D.~Katz, J.~Kenney, and O.~Brock, ``How can robots succeed in unstructured
  environments,'' in \emph{In Workshop on Robot Manipulation: Intelligence in
  Human Environments at Robotics: Science and Systems}.\hskip 1em plus 0.5em
  minus 0.4em\relax Citeseer, 2008.

\bibitem{dogar2011framework}
M.~Dogar and S.~Srinivasa, ``A framework for push-grasping in clutter,''
  \emph{Robotics: Science and systems VII}, vol.~1, 2011.

\bibitem{danielczuk2019segmenting}
M.~Danielczuk, M.~Matl, S.~Gupta, A.~Li, A.~Lee, J.~Mahler, and K.~Goldberg,
  ``Segmenting unknown 3d objects from real depth images using mask r-cnn
  trained on synthetic data,'' in \emph{2019 International Conference on
  Robotics and Automation (ICRA)}.\hskip 1em plus 0.5em minus 0.4em\relax IEEE,
  2019, pp. 7283--7290.

\bibitem{Koch2015SiameseNN}
G.~Koch, R.~Zemel, and R.~Salakhutdinov, ``Siamese neural networks for one-shot
  image recognition,'' 2015.

\bibitem{mahler2017dex2}
J.~Mahler, J.~Liang, S.~Niyaz, M.~Laskey, R.~Doan, X.~Liu, J.~A. Ojea, and
  K.~Goldberg, ``Dex-net 2.0: Deep learning to plan robust grasps with
  synthetic point clouds and analytic grasp metrics,'' \emph{arXiv preprint
  arXiv:1703.09312}, 2017.

\bibitem{pineau}
J.~Pineau, ``Tractable planning under uncertainty: Exploiting structure,''
  Ph.D. dissertation, Robotics Institute, Carnegie Mellon University, 2004.

\bibitem{kulkarni2016hierarchical}
T.~D. Kulkarni, K.~Narasimhan, A.~Saeedi, and J.~Tenenbaum, ``Hierarchical deep
  reinforcement learning: Integrating temporal abstraction and intrinsic
  motivation,'' \emph{Advances in neural information processing systems},
  vol.~29, pp. 3675--3683, 2016.

\bibitem{7759839}
J.~K. Li, D.~Hsu, and W.~S. Lee, ``Act to see and see to act: Pomdp planning
  for objects search in clutter,'' in \emph{2016 IEEE/RSJ International
  Conference on Intelligent Robots and Systems (IROS)}, 2016, pp. 5701--5707.

\bibitem{mahler2018dex3}
J.~Mahler, M.~Matl, X.~Liu, A.~Li, D.~Gealy, and K.~Goldberg, ``Dex-net 3.0:
  Computing robust vacuum suction grasp targets in point clouds using a new
  analytic model and deep learning,'' in \emph{2018 IEEE International
  Conference on robotics and automation (ICRA)}.\hskip 1em plus 0.5em minus
  0.4em\relax IEEE, 2018, pp. 5620--5627.

\bibitem{LinPush}
M.~Danielczuk, J.~Mahler, C.~Correa, and K.~Goldberg, ``Linear push policies to
  increase grasp access for robot bin picking,'' in \emph{2018 IEEE 14th
  International Conference on Automation Science and Engineering (CASE)}.\hskip
  1em plus 0.5em minus 0.4em\relax IEEE, 2018, pp. 1249--1256.

\bibitem{DanielczukAVG20}
M.~Danielczuk, A.~Angelova, V.~Vanhoucke, and K.~Goldberg, ``X-ray: Mechanical
  search for an occluded object by minimizing support of learned occupancy
  distributions,'' in \emph{{IEEE/RSJ} International Conference on Intelligent
  Robots and Systems, {IROS} 2020, Las Vegas, NV, USA, October 24, 2020 -
  January 24, 2021}.\hskip 1em plus 0.5em minus 0.4em\relax {IEEE}, 2020, pp.
  9577--9584.

\bibitem{kurenkov2020visuomotor}
A.~Kurenkov, J.~Taglic, R.~Kulkarni, M.~Dominguez-Kuhne, A.~Garg,
  R.~Mart{\'\i}n-Mart{\'\i}n, and S.~Savarese, ``Visuomotor mechanical search:
  Learning to retrieve target objects in clutter,'' in \emph{2020 IEEE/RSJ
  International Conference on Intelligent Robots and Systems (IROS)}.\hskip 1em
  plus 0.5em minus 0.4em\relax IEEE, 2020, pp. 8408--8414.

\bibitem{modularrl21}
I.~{Sarantopoulos}, M.~{Kiatos}, Z.~{Doulgeri}, and S.~{Malassiotis}, ``Total
  singulation with modular reinforcement learning,'' \emph{IEEE Robotics and
  Automation Letters}, pp. 1--1, 2021.

\bibitem{9197101}
T.~Novkovic, R.~Pautrat, F.~Furrer, M.~Breyer, R.~Siegwart, and J.~Nieto,
  ``Object finding in cluttered scenes using interactive perception,'' in
  \emph{2020 IEEE International Conference on Robotics and Automation (ICRA)},
  2020, pp. 8338--8344.

\bibitem{pan2020decision}
Z.~Pan and K.~Hauser, ``Decision making in joint push-grasp action space for
  large-scale object sorting,'' \emph{arXiv preprint arXiv:2010.10064}, 2020.

\bibitem{zeng2018learning}
A.~Zeng, S.~Song, S.~Welker, J.~Lee, A.~Rodriguez, and T.~Funkhouser,
  ``Learning synergies between pushing and grasping with self-supervised deep
  reinforcement learning,'' in \emph{2018 IEEE/RSJ International Conference on
  Intelligent Robots and Systems (IROS)}.\hskip 1em plus 0.5em minus
  0.4em\relax IEEE, 2018, pp. 4238--4245.

\bibitem{8967899}
Y.~Deng, X.~Guo, Y.~Wei, K.~Lu, B.~Fang, D.~Guo, H.~Liu, and F.~Sun, ``Deep
  reinforcement learning for robotic pushing and picking in cluttered
  environment,'' in \emph{2019 IEEE/RSJ International Conference on Intelligent
  Robots and Systems (IROS)}, 2019, pp. 619--626.

\bibitem{berscheid2019robot}
L.~Berscheid, P.~Mei{\ss}ner, and T.~Kr{\"o}ger, ``Robot learning of shifting
  objects for grasping in cluttered environments,'' in \emph{2019 IEEE/RSJ
  International Conference on Intelligent Robots and Systems (IROS)}.\hskip 1em
  plus 0.5em minus 0.4em\relax IEEE, 2019, pp. 612--618.

\bibitem{bcai}
\BIBentryALTinterwordspacing
Z.~Feldman, H.~Ziesche, N.~A. Vien, and D.~D. Castro, ``A hybrid approach for
  learning to shift and grasp with elaborate motion primitives,'' \emph{CoRR},
  vol. abs/2111.01510, 2021. [Online]. Available:
  \url{https://arxiv.org/abs/2111.01510}
\BIBentrySTDinterwordspacing

\bibitem{VienT15}
N.~A. Vien and M.~Toussaint, ``Hierarchical monte-carlo planning,'' in
  \emph{Proceedings of the Twenty-Ninth {AAAI} Conference on Artificial
  Intelligence, January 25-30, 2015, Austin, Texas, {USA}}, B.~Bonet and
  S.~Koenig, Eds.\hskip 1em plus 0.5em minus 0.4em\relax {AAAI} Press, 2015,
  pp. 3613--3619.

\bibitem{white1976procedures}
C.~C. White, ``Procedures for the solution of a finite-horizon, partially
  observed, semi-markov optimization problem,'' \emph{Operations Research},
  vol.~24, no.~2, pp. 348--358, 1976.

\bibitem{HaarnojaZAL18}
T.~Haarnoja, A.~Zhou, P.~Abbeel, and S.~Levine, ``Soft actor-critic: Off-policy
  maximum entropy deep reinforcement learning with a stochastic actor,'' in
  \emph{Proceedings of the 35th International Conference on Machine Learning,
  {ICML} 2018, Stockholmsm{\"{a}}ssan, Stockholm, Sweden, July 10-15, 2018},
  ser. Proceedings of Machine Learning Research, J.~G. Dy and A.~Krause, Eds.,
  vol.~80.\hskip 1em plus 0.5em minus 0.4em\relax {PMLR}, 2018, pp. 1856--1865.

\bibitem{mnih2015human}
V.~Mnih, K.~Kavukcuoglu, D.~Silver, A.~A. Rusu, J.~Veness, M.~G. Bellemare,
  A.~Graves, M.~Riedmiller, A.~K. Fidjeland, G.~Ostrovski, \emph{et~al.},
  ``Human-level control through deep reinforcement learning,'' \emph{nature},
  vol. 518, no. 7540, pp. 529--533, 2015.

\bibitem{coumans2017pybullet}
E.~Coumans and Y.~Bai, ``Pybullet, a python module for physics simulation in
  robotics, games and machine learning,'' 2017.

\bibitem{calli2017ycb}
B.~Calli, A.~Singh, J.~Bruce, A.~Walsman, K.~Konolige, S.~Srinivasa, P.~Abbeel,
  and A.~M. Dollar, ``Yale-cmu-berkeley dataset for robotic manipulation
  research,'' \emph{The International Journal of Robotics Research}, vol.~36,
  no.~3, pp. 261--268, 2017.

\bibitem{keselman2017realsense}
L.~Keselman, J.~Iselin~Woodfill, A.~Grunnet-Jepsen, and A.~Bhowmik, ``Intel
  realsense stereoscopic depth cameras,'' in \emph{Proceedings of the IEEE
  Conference on Computer Vision and Pattern Recognition Workshops}, 2017, pp.
  1--10.

\end{thebibliography}

\end{document}